%% file: main.tex
\documentclass{article}

\usepackage[preprint]{corl_2026} % Uncomment for pre-prints (e.g., arxiv); This is like ``final'', but will remove the CORL footnote.
\usepackage{amsmath}
\usepackage{amsfonts}
\usepackage{amssymb}
\usepackage{booktabs}
\usepackage{caption}
\usepackage{graphicx}
\usepackage{xspace}
\usepackage{multirow}
\usepackage{makecell}
\usepackage{marvosym}

\title{CLEAR: Closed-Loop Reinforcement Learning at Scale
for End-to-End Autonomous Driving}

\author{
Yunxiao Shi$^\text{\Letter}$~~~
Hong Cai$^\text{\Letter}$~~~
Mohammad Ghavamzadeh~~~
Fatih Porikli~~~
\smallskip
\\[1pt]
{Qualcomm AI Research}
\\[1pt]
\smallskip
{\tt\footnotesize\{yunxshi, hongcai\}@qti.qualcomm.com}\\
{\footnotesize \it Qualcomm AI Research is an initiative of Qualcomm Technologies, Inc}
}

\newcommand{\ours}{CLEAR\xspace}

\begin{document}
\maketitle
%===============================================================================

\input{secs/0_abstract}

\keywords{End-to-End Autonomous Driving, Reinforcement Learning, VLA} 

%===============================================================================

%===============================================================================

% \section{Citations}
% \label{sec:citations}

%	Citations can be made using either \textbackslash citep\{\} or \textbackslash citet\{\}, depending from the appropriateness. To avoid the citation moving to the next line, it is often a good practice to replace the space before with a tilde (\~{}) character.
%	Example 1: ``CoRL is the best conference ever~\citep{Gauss1857}.''
%	Example 2: ``\citet{Lagrange1788} proved, both theoretically and numerically, that CoRL is the best conference ever.''

\input{secs/1_intro}
\input{secs/2_related}
\input{secs/3_method}
\input{secs/4_experiment}

\input{secs/5_limitation}

\input{secs/6_conclusion}

\clearpage
% The acknowledgments are automatically included only in the final and preprint versions of the paper.
% \acknowledgments{If a paper is accepted, the final camera-ready version will (and probably should) include acknowledgments. All acknowledgments go at the end of the paper, including thanks to reviewers who gave useful comments, to colleagues who contributed to the ideas, and to funding agencies and corporate sponsors that provided financial support.}

%===============================================================================

\bibliography{main}  % .bib

\end{document}

%% file: secs/0_abstract.tex
\begin{abstract}
    End-to-end autonomous driving (E2E-AD) aims to directly map raw sensor information to driving actions. Recently, with the rapid advancement of multi-modal large language models (MLLMs), researchers have proposed the paradigm of Vision-Language-Action (VLA) models for E2E-AD, where it seeks to integrate visual perception, language understanding and action prediction within a single policy. However, existing VLA-based policies largely adopts imitation learning, where it only learns to drive by optimizing distance-based metrics \textit{w.r.t.} logged expert trajectories. Such distribution shift between open-loop training and closed-loop inference leads to suboptimal performance in closed-loop planning. To close this gap, we present \ours, a system that enables closed-loop training using Reinforcement Learning (RL) at scale for E2E-AD. We propose to learn a novel residual waypoint policy around the waypoint prior from pretrained VLA policies, effectively harnessing the knowledge within. On another front, one of the key challenges to scale up RL for vision-based policies is the number of parallel simulation environments since RL is data hungry. To that end, we design a heterogeneous pipeline that places the simulator and the VLA learner on distinct compute groups, which allows us to dramatically increase the number of simulation environments running in parallel while avoiding resource contention and maintaining training stability. We show that with a simple reward, \ours significantly outperforms previous methods and sets new state-of-the-art performance on the challenging benchmarks of CARLA longest6 v2 and Bench2Drive.
\end{abstract}

%% file: secs/1_intro.tex
\section{Introduction}
\label{sec:intro}

End-to-end autonomous driving (E2E-AD)~\cite{chen2024end,hu2023planning,jiang2023vad,chen2024vadv2,li2025navigationguidedsparsescenerepresentation} has emerged as a powerful alternative to traditional modular approaches~\cite{hawke2020urban,chitta2021neat,chitta2022transfuser,jia2023think,shao2023safety}. Recently, an explosion of research has appeared on using LLMs/VLMs~\cite{achiam2023gpt,team2023gemini,qwen, chen2024internvl,comanici2025gemini,singh2025openai} to bridge the reasoning gap for autonomous driving, which paved the way for Vision-Langage-Action (VLA) models for E2E-AD. VLAs aim to be able to interpret spontaneous, high-level driving commands, perform robust reasoning under rare scenarios, and have noise-tolerant control in dynamic regions~\cite{jiang2025survey}. Within a short period of time, VLA agents have demonstrated strong results on various benchmarks~\cite{caesar2020nuscenes,caesar2021nuplan,sun2020scalability,dauner2024navsim,jia2024bench2drive,cao2025pseudo}. 

Existing VLA-based policies~\cite{hwang2024emma,xing2025openemma,fu2025orion,renz2025simlingo} largely adopts imitation learning (IL) as the primary approach, \textit{i.e.}, the learning objective typically is only to optimize certain distance metrics (\textit{e.g.}, $L_2$ error) \textit{w.r.t.} ground-truth expert trajectories. The benefit of IL is that it can scale with data. However, the distribution shift between open-loop training and closed-loop inference often leads to compromised policies, exhibiting suboptimal performance in closed-loop planning. More recently methods like~\cite{li2026recogdrive} explore using Reinforcement Learning (RL)~\cite{sutton1998reinforcement} to improve planning but only under an open-loop setup, still falling short when tested in a closed-loop environment.

%Hence, it is imperative to go beyond simple behavior cloning, and equip driving agents with the ability to learn complex driving dynamics in an interactive manner. 

%\vspace{-2pt}
A recent line of research~\cite{renz2022plant,wu2022trajectory,li2024think2drive,jaeger2025carl} explores closed-loop training using reinforcement learning. However, they only consider the task of privileged planning, where perfect perception ground truths and map information are required as inputs. However, such information is not always available at training time and this does not reflect real-world deployment where there can be perception errors.

In this work, we propose \ours, a system that effectively enables and efficiently performs closed-loop RL finetuning of VLA policies at scale for E2E-AD. Given the size of VLA models, carrying out RL finetuning from scratch is not feasible in terms of the amount of compute needed. Therefore, we first pretrain \ours on a large set of expert trajectories using imitation learning, equipping it with baseline driving capabilities. Then, we learn a novel residual waypoint policy to correct the waypoints predictions from the pretrained VLA, which we eventually convert to low-level control for driving. We optimize the policy with Proximal Policy Optimization (PPO)~\cite{schulman2017proximal}, given its stable convergence and high asymptotic performance. In addition, we design a heterogeneous pipeline which allows us to significantly increase the number of parallel simulation environments, which is critical for effective RL. A summary of the design of \ours is shown in Fig.~\ref{fig:clear}.

%% file: secs/2_related.tex
\section{Related Works}
\label{sec:related}
\vspace{-3pt}
\subsection{End-to-End Autonomous Driving}
\vspace{-3pt}
Early E2E-AD systems integrates perception, prediction and planning tasks in a single stack and impose intermediate supervision~\cite{hu2023planning,chen2024vadv2,jiang2023vad}. Later, numerous research efforts~\cite{huang2024drivemm,fu2024drive,liu2023mtd,ma2024dolphins,paul2024lego,wang2024drive,wang2023drivemlm} explore the use of LLM/VLM for E2E-AD. GPT-Driver~\cite{mao2023gpt} and EMMA~\cite{hwang2024emma} cast trajectory prediction as a language prediction problem. DriveVLM~\cite{tian2024drivevlm} and Senna~\cite{jiang2024senna} propose dual systems where the VLM generates low-frequency trajectories or driving commands to guide the end-to-end model to produce the final trajectory. ORION~\cite{fu2025orion} couples transformer memory and LLM to summarize the history to output the next trajectory. SimLingo~\cite{renz2025simlingo} proposes action dreaming to better align language and driving action. However, all these methods are imitation learning driven, which is insufficient for closed-loop planning.

%Later approaches explores modeling interactions between ego-vehicle and other traffic agents~\cite{zheng2024genad,chen2024ppad,zhang2024graphad}, sparse architectures for better efficiency~\cite{sun2025sparsedrive,zhu2025sparsead,li2025navigationguidedsparsescenerepresentation}, and eliminating dense BEV features for better scalability~\cite{weng2024drive,jia2025drivetransformer}. More attempts~\cite{guo2025end,li2025enhancing,lu2024activead,li2024does} explore self-supervised learning or weakly supervised learning to reduce the reliance 3D perception annotations. Overall, these systems are vulnerable to rare scenarios and struggle with long-tail generalization.

\subsection{Closed-Loop Reinforcement Learning}
\vspace{-3pt}
Reinforcement Learning~\cite{sutton1998reinforcement,ouyang2022training} is a promising venue for closed-loop training. Efforts on privileged planning~\cite{renz2022plant,jaeger2025carl} have explored using RL for driving, where perfect ground-truth (\textit{e.g.}, 3D boxes~\cite{renz2022plant}, BEV segmentation maps~\cite{jaeger2025carl}) and global visibility of the environment (\textit{e.g.} HD/SD maps) are needed as input information. For E2E-AD, RAD~\cite{gao2026rad} utilizes 3D Gaussian Splatting for scenario generation and performs RL within. TrajHF~\cite{li2025finetuning} and Gen-Drive~\cite{huang2025gen} aligns trajectory generation with safety constraints and human preferences. AlphaDrive~\cite{jiang2025alphadrive} and AutoVLA~\cite{zhou2026autovla} adopts GRPO~\cite{shao2024deepseekmath} to enhance planning performance and make reasoning more efficient. However, recent efforts only investigated RL for E2E-AD at a limited scale. In this work, we propose a system that enables RL at scale for E2E-AD and unlock new levels of planning performance and robustness.

%% file: secs/3_method.tex
\section{Method}
\label{sec:method}

We first provide an overview of E2E-AD in Section~\ref{sec:overview}. In Section~\ref{sec:openloop}, we briefly go over open-loop pretraining. We detail our RL finetuning framework in Section~\ref{sec:closedloop}. We describe our heterogeneous pipeline in Section~\ref{sec:hetero}.

\begin{figure*}[h]
    \centering
    \includegraphics[width=\linewidth]{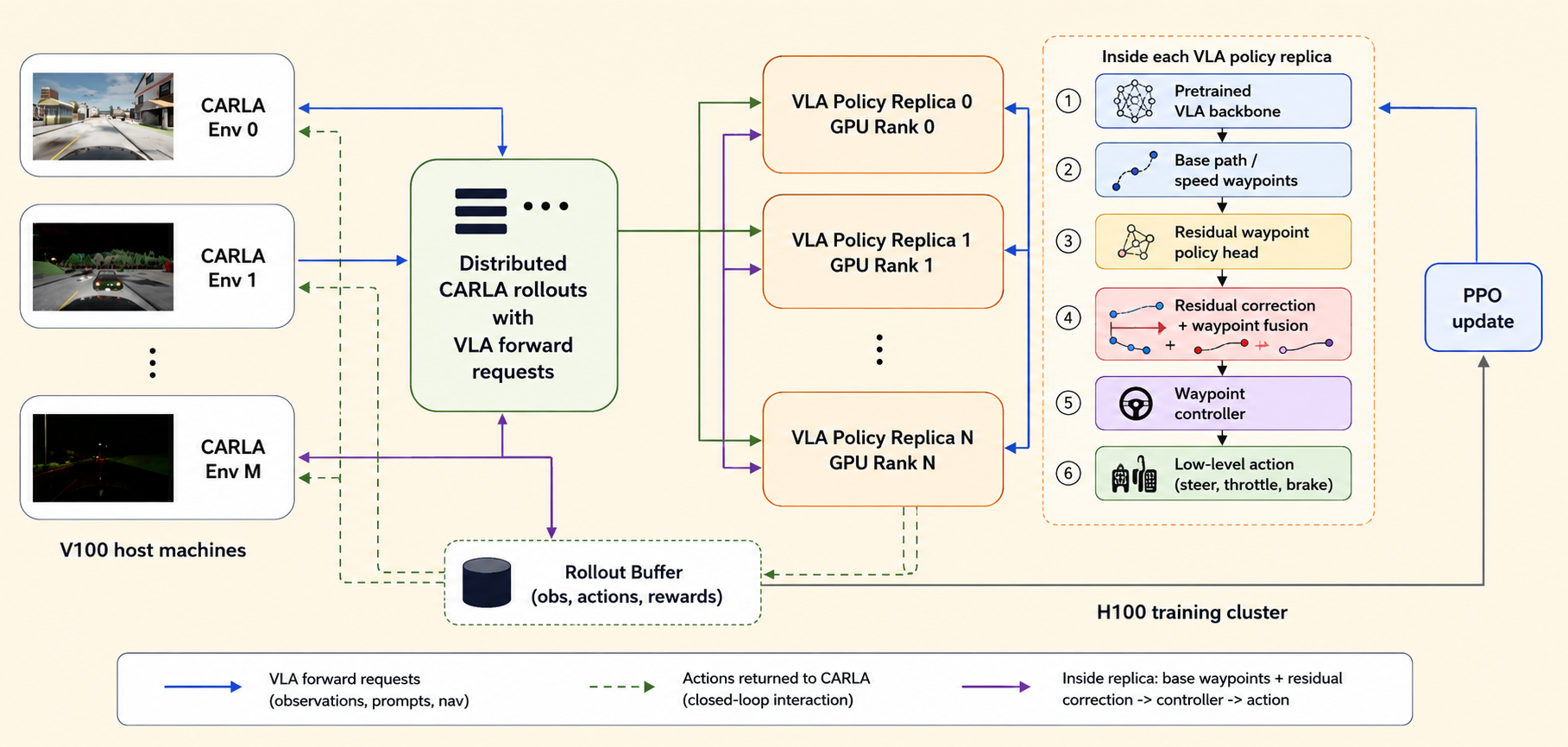}
    \caption{\small \ours system pipeline. We conduct closed-loop RL finetuning on top of VLAs pretrained using imitation learning, and learn a residual policy around the pretrained waypoints prior. With our heterogeneous finetuning pipeline, we scale up the number of simulators~\cite{dosovitskiy2017carla} running in parallel and PPO~\cite{schulman2017proximal,wijmans2019dd} updates dramatically.}
    \label{fig:clear}
\end{figure*}

\subsection{Problem Setup}
\label{sec:overview}

We study closed-loop finetuning of a vision-language-action model for end-to-end autonomous driving. At each time step $t$, the ego vehicle receives multi-camera images $\{I_t^i\}_{i=1}^N$ where $N$ is the number of cameras, ego speed $v_t$, navigation information $n_t$ (\textit{e.g.} target GPS points or high-level driving commands), and a task for language prompt $l_t$. The goal is to learn a driving policy that maps these observations to robust vehicle behavior. Formally, we can formulate this as a Partially Observable Markov Decision Process (POMDP),
\begin{equation}
    \mathcal{M} = (\mathcal{O}, \mathcal{A}, \mathcal{P}, r, \gamma),
\end{equation}
where $o_t = \{I_t, v_t, n_t, l_t\}\in\mathcal{O}$ is the observation, $a_t\in\mathcal{A}$ is the policy action, $\mathcal{P}$ demotes simulator transition dynamics, $r_t$ is the reward and $\gamma$ is the discount factor. A straight-forward way is to enable a direct-control policy that would output low-level driving commands
\begin{equation}
    a_t = (\delta_t, \tau_t, b_t),
\end{equation}
where $\delta_t\in[0, 2\pi]$, $\tau_t\in[0, 1]$ and $b_t\in[0, 1]$ denote steering, throttle and brake. However, most (if not all) existing VLA policies~\cite{renz2025simlingo,hwang2024emma,xing2025openemma,fu2025orion} are pretrained on a large amount of expert trajectories, \textit{i.e.}, waypoints annotations, a direct-control policy would be poorly aligned with the waypoint-based action representation learned during the pretraining stage. Therefore, we propose a novel finetuning strategy in a residual waypoint space, where the pretrained VLA predicts a base path, and the RL policy learns bounded residual corrections to such plan. Given the CARLA simulator expects driving commands, a deterministic controller is used to map the resulting waypoint command to low-level CARLA control.

Our method \ours has two stages. First, we utilize open-loop VLA pretraining to learn a vision-langauge-conditioned waypoint prior. Second, we finetune the VLA policy in a closed-loop simulator (CARLA) with PPO~\cite{schulman2017proximal}. In addition, to scale up the number of parallel simulation environments which is critical for data-hungry RL, we design a heterogeneous pipeline where the server (simulator) and client (VLA learner) are placed on different compute platforms, mitigating the issue of resource contention and enabling effective scaling.

\subsection{VLA Open-Loop Pretraining}
\label{sec:openloop}

We use a SimLingo-style VLA~\cite{renz2025simlingo} as our open-loop driving prior, where we swap out the original InternVL2-1B~\cite{chen2024internvl} with InternVL3-1B~\cite{zhu2025internvl3} given its overall better performance. InternVL3-1B consists of an InternViT-300M-448px vision encoder~\cite{chen2024internvl}, and a Qwen2.5-0.5B~\cite{Bai2025qwen25} large language model (LLM). We briefly describe the various aspects involved in VLA pretraining below. 

Denote
\begin{equation}
    e_{I} = f_{\text{vis}}(\{I_t^i\}_{i=1}^N),\quad e_{\text{nav}} = f_{\text{nav}}(n_t),\quad e_{l} = f_{\text{tok}}(l_t)
\end{equation}
as the image, navigation, and language prompt embeddings. Here we use $N=1$ and convert $I_t$ to image tiles as input following~\cite{renz2025simlingo}. The interleaved LLM input is
\begin{equation}
    e_{\text{LLM}} = \text{TI}(e_{\text{vis}}, e_{\text{nav}}, e_l),
\end{equation}
where $\text{TI}(\cdot)$ denotes token interleaving. The VLA backbone produces driving features and action-query outputs, followed by generating path and speed waypoints~\cite{renz2025simlingo}
\begin{align}
    &[h_t, O_t^p, O_t^w] = F_{\theta}(e_{\text{LLM}}, q_p, q_w),\\
    &\hat{\mathbf{P}}_t = \psi_p(O_t^p) = \{\hat{\mathbf{p}}_t^1, \dots, \hat{\mathbf{p}}_t^{N_p}\},\quad \hat{\mathbf{p}}_t^i = (\hat{x}_{t,i}^p, \hat{y}_{t,i}^p)\\
    &\hat{\mathbf{W}}_t = \psi_w(O_t^w) = \{\hat{\mathbf{w}}_t^1, \dots, \hat{\mathbf{w}}_t^{N_w}\},\quad \hat{\mathbf{w}}_t^i = (\hat{x}_{t,i}^w, \hat{y}_{t,i}^w),
\end{align}
where $F_{\theta}$ denotes the LLM. $q_p$ and $q_w$ are the corresponding action queries~\cite{renz2025simlingo}. $\psi_p$ and $\psi_w$ are two MLPs that decodes to $\hat{\mathbf{P}}_t$ and $\hat{\mathbf{W}}_t$ which are predicted path and speed waypoints. $N_p$ and $N_w$ are the number of corresponding waypoints. During the pretraining stage, $\hat{\mathbf{P}}_t$ and $\hat{\mathbf{W}}_t$ are supervised with ground-truth expert labels $\mathbf{P}^{*}_t$ and $\mathbf{W}^{*}_t$. 

%We extract the visual tokens as $v_e = \delta_{\downarrow}([\text{V}_e(I_i)]_{i=0}^{N_i}\in\mathbb{R}^{(N_iM)\times d}$, where $I_i$ are the input image tiles, $V_e$ is the vision encoder, $\delta_{\downarrow}(\cdot)$ is the downsampling operator~\cite{shi2016real,chen2024internvl,renz2025simlingo} which downsamples the number of tokens to $M=256$. Here we set the number of image tiles $N_i=2$ and $d$ is the embedding dimension. Denote $L_e$ and $Nav_e$ the embeddings of the rest input modalities, we generate waypoints prediction as 
%\begin{equation}
%    \{W, P\} = \text{MLP}(\text{LLM}([\{v_e, L_e, Nav_e\}, q_w, q_p])),
%\end{equation} 
%where $q$ are the action queries introduced %in~\cite{renz2025simlingo}.

\subsection{Closed-Loop RL Finetuning}
\label{sec:closedloop}

Unlike prior privileged RL driving policies~\cite{wu2022trajectory,jaeger2025carl} that directly output low-level controls, our closed-loop finetuning operates in the waypoint space of the pretrained VLA. We treat the open-loop VLA as a trajectory prior and learn a residual policy around it. Concretely, at each time step $t$, given
\begin{equation}
    (h_t, \hat{\mathbf{P}}_t, \hat{\mathbf{W}}_t) = \mathcal{F}_{\theta_0}(I_t, v_t, n_t, l_t),
\end{equation}
where $\mathcal{F}$ denotes the pretrained VLA and  $\theta_0$ its parameters, $h_t$ is the driving feature. We first define $m$ fixed longitudinal anchor locations in the ego frame,
\begin{equation}
    \mathbf{x}^{\text{anc}} = [x^1, \dots, x^m].
\end{equation}
These anchors are chosen to lie within the reliable prediction horizon of the predicted waypoints. Given with the pretrained geometric path waypoints $\hat{\mathbf{P}}_t$, we extract base lateral offsets at at each anchor by interpolation,
\begin{align}
    \hat{\mathbf{y}}_t &= \text{Interp}(\hat{\mathbf{P}_t}, x^k),\,k = 1, \dots, m,  \\
    &= [\hat{y}^1, \dots, \hat{y}_m].
\end{align}
With the temporal speed waypoints $\hat{\mathbf{W}}_t$, we estimate a base target speed,
\begin{equation}
    \hat{v}_t = \phi(\hat{\mathbf{W}}_t),
\end{equation}
where $\phi$ computes speed from the displacements of the predicted temporal speed waypoints. The RL policy conditions on both the driving feature and the base plan, and samples a residual waypoint action,
\begin{align}
    \hat{u}_t^{\text{wp}} &= [\hat{y}_t^1, \dots, \hat{y}_t^m, \hat{v}_t], \\
    z_t &= \text{MLP}([h_t, \hat{u}_t^{\text{wp}}]),\\
    a_t &= [\Delta y_t^1, \dots, \Delta y_t^m, \Delta v_t]\sim \pi_\theta(\cdot | z_t).
\end{align}
And we obtain the waypoint command by adding the residual to the base plan,
\begin{align}
    y_t^i &= \Gamma(\hat{y}_t^i + \Delta y_t^i, -y_{\text{min}}, y_\text{\text{max}}),\\ v_t^{\text{cmd}} &= \Gamma(\hat{v}_t + \Delta v_t, v_{\text{min}}, v_{\text{max}}),\quad i = 1, \dots, m,.
\end{align}
where $\Gamma$ is the clipping function. Therefore, the final executed waypoint command is
\begin{equation}
    u_t^{\text{wp}} = [(x^1, y_t^1), \dots, (x^m, y_t^m), v_t^{\text{cmd}}],
\end{equation}
which is passed to the driving controller to map to low-level vehicle control,
\begin{equation}
    c_t = g(u_t^{\text{wp}}, s_t) = (\delta_t, \eta_t),
\end{equation}
where $\eta_t\in[-1, 1]$ merges throttle $\tau_t$ and brake $b_t$ in a single action dimension since they are mutually exclusive. We note that the controller is treated as part of the environment transition
\begin{equation}
    \{o_{t+1}, r_t, d_t\} \sim \mathcal{E}(o_t, g(u_t^{\text{wp}}, s_t)),
\end{equation}
where $g$ is the controller, $d_t$ denotes the termination indicator. Therefore, no gradient through the controller is required. PPO optimizes the probability of the sampled residual waypoint action $a_t$, while the controller only converts the final waypoint action into executable control.
In addition, we freeze both the vision encoder and LLM in $\mathcal{F}_{\theta_0}$ entirely and also disable the auto-regressive language prediction path during the RL finetuning stage, due to limitation of compute resources.

%We make use of the CARLA driving simulator~\cite{dosovitskiy2017carla} release 0.9.15 to conduct closed-loop RL finetuning. Given the compute limitations, we freeze the vision encoder and LLM in the pretrained VLA entirely. Since CARLA environments expect driving actions, we replace the waypoints head with a small action head, which predicts steering $s\in[0, 1]$, and throttle/brake which are merged into a single action dimension $tb\in[-1, 1]$ since they are mutually exclusive. For reward design, we adopt the discounted Route Completion (RC)~\cite{jaeger2025carl,carlaleaderboard} 
%\begin{equation}
%    r_t = RC_t\cdot(\Pi p_t)-T   
%\end{equation}
%as our reward, where $t$ denotes each time step, $p_t\in[0, 1]$ is the soft penalty factor, and $T=\{0, 1\}$ is the hard penalty (\textit{e.g.}, collision, red light/stop sign infraction, etc) which terminates an episode. We utilize DD-PPO~\cite{wijmans2019dd} to scale up PPO updates under a distributed setting. To further improve efficiency, we also disable the auto-regressive language prediction path in the pretrained VLA during the RL finetuning stage.

\subsection{Heterogeneous Finetuning Pipeline}
\label{sec:hetero}

One of the key challenges of performing RL for vision-based policies is to scale up the number of simulation environments running in parallel. However, the GPU rendering stack of CARLA is known for its instability and brittleness~\cite{jia2024bench2drive,carlaissues}. We observed frequent CARLA server crashes during RL training especially in containerized environments (\textit{e.g.}, docker). In addition, a single CARLA server takes about 6GB of GPU memory to initialize and run, which carves into the total memory budget (\textit{e.g.}, a Hopper H100 GPU has a total of 80GB GPU memory) that the VLA also need to consume from, and itself is memory-intensive. Such facts lead to the problem of resource contention, which stifles the scaling of parallel CARLA environments, and also contributes to training instability. 

To address this, we design a heterogeneous finetuning pipeline that allows us to dramatically increase the number of parallel CARLA simulators. Unlike latest VLA models that typically need high-end GPUs (\textit{e.g.}, B200/H100/...) to run efficiently, CARLA (and the underlying Unreal Engine~\cite{engine2018unreal}) does not have such stringent requirement and can run on older-generation (\textit{e.g.}, Tesla V100) or even consumer-grade GPUs. Besides, we observe that CARLA works more smoothly with lower NVIDIA drivers (\textit{e.g.}, version $\leq 535$). This prompted us to place the simulator and learner on distinct compute infrastructures to unlock scaling and resolve resource contention. As shown in Fig.~\ref{fig:clear}, we launch CARLA servers on standalone host machines, while putting the VLA policies on high-performance H100 clusters. To establish communication between server and client, we construct SSH tunnels~\cite{rfc4254} to facilitate learning. With such heterogeneous design, we are able to launch 64 CARLA environments in parallel, and scale to 100M samples with a total-batch/mini-batch of 16384/4096 for PPO updates, which lead to superior results.

\subsection{Training Objectives}

During the open-loop pretraining stage, we supervise predicted path and speed waypoints with smooth $L_1$ loss, along with language supervision for the task of VQA. We refer readers to~\cite{renz2025simlingo} for such details.

For closed-loop finetuning, we optimize the expected closed-loop return,
\begin{equation}
    J(\theta) = \mathbb{E}_{\pi_\theta}\Big[\sum_{t=0}^T\gamma^tr_t\Big],\quad r_t = RC_t\cdot(\prod p_t) - P,
\end{equation}
where $\gamma^t$ is the discount factor, $r_t$ is the reward based on discounted Route Completion (RC) from~\cite{jaeger2025carl,carlaleaderboard}. $p_t\in[0, 1]$ is the soft penalty factor and $P$ is the hard penalty (\textit{e.g.}, collision, red light/stop sign infraction, etc) which terminates an episode.

%% file: secs/4_experiment.tex
\section{Experiments}
\label{sec:experiment}
\vspace{-3pt}
\subsection{Implementation Details}

We leverage the dataset curated by~\cite{renz2025simlingo} which totals about 3.1 million samples at 4 fps for open-loop pretraining. We choose InternVL3-1B~\cite{zhu2025internvl3} to be our VLA backbone. AdamW~\cite{loshchilov2017decoupled} is used as the optimizer with a weight decay of 0.1, and an initial learning rate of 3e-5 with a cosine annealing schedule. We pretrain the VLA on 8 H100 GPUs for 14 epochs with a per-GPU batch size of 12 (global batch size of 96). LLM portion of the VLA is finetuned using LoRA~\cite{hu2022lora} with the same config as in~\cite{renz2025simlingo}.

For closed-loop finetuning, we make use of the procedurally generated routes from a suite of parallel CARLA environments by~\cite{jaeger2025carl}. We leverage 32 Tesla V100 (32GB) GPUs for launching CARLA servers and assign 2 instances per GPU, resulting in 64 CARLA simulation environments running in parallel. We observe such that setup gives the most stable training behavior while enabling scaling. Pretrained VLA policy is placed on 8 H100 GPUs same as in the pretraining stage. We set the anchor points horizon $m=4$. Given the distributed setting, we make use of DD-PPO~\cite{wijmans2019dd} for PPO updates. We construct SSH tunnels~\cite{rfc4254} to establish communication between the simulation environment and learner. To ensure maximum uptime, we utilize autossh. With our setup, we scale to 100M samples for RL training with a total batch size/mini-batch size of 16384/4096. Training roughly takes 6 days to conclude.

\begin{figure*}[t]
    \centering
    \includegraphics[width=\textwidth]{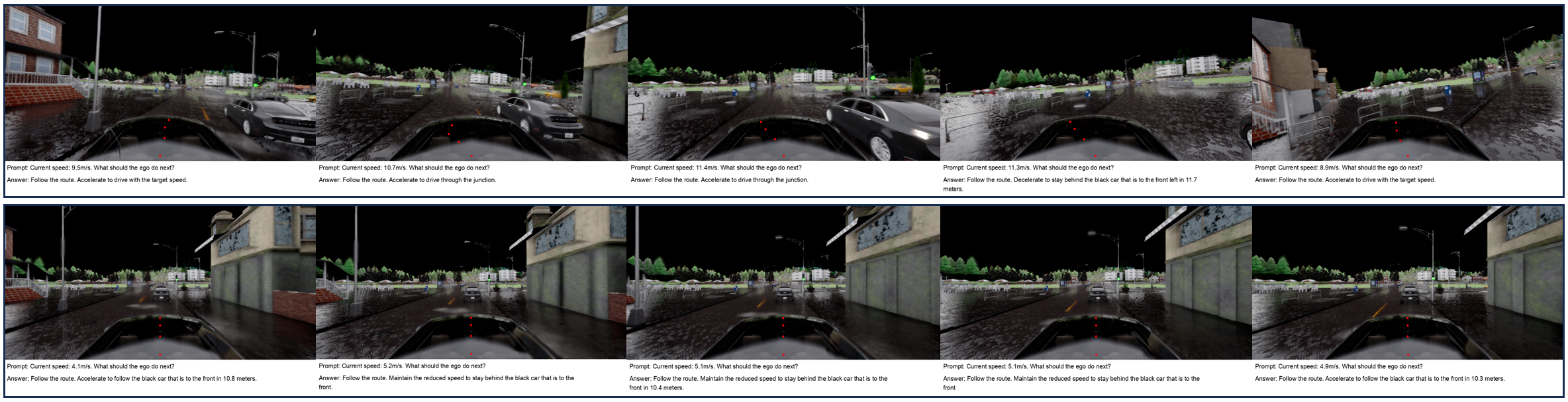}
    \vspace{-17pt}
    \caption{\small Qualitative predictions of \ours on the CARLA longest6 v2~\cite{jaeger2025carl,carlaleaderboard} benchmark. We visualize the pretrained waypoints horizon (first row) vs. our residual corrected ones (second row). One can see that pretrained policy gave the wrong prediction and led to a crash, whereas our policy followed the traffic smooothly. Best viewed in color and zoomed in.}
    \label{fig:demo-clear}
    \vspace{-8pt}
\end{figure*}

\begin{table*}[t]
  \small
  \centering
  \caption{\small Closed-loop planning performance on the CARLA longest6 v2 benchmark~\cite{jaeger2025carl,carlaleaderboard}. IL stands for ``Imitation Learning''. We note that all methods here are non-privileged planners.}
  \vspace{-5pt}
  \resizebox{0.65\textwidth}{!}{
  \begin{tabular}{ccccc}
    \toprule
    Method  & Mode & DS$\uparrow$ & SR (\%)$\uparrow$ & RC (\%)$\uparrow$ \\
    \midrule
    UniAD~\cite{hu2023planning} & IL &  4.64	& 0.00	& 8.78\\
    VAD~\cite{jiang2023vad}  & IL & 3.95 &	0.00	& 6.44\\
    SSR~\cite{li2025navigationguidedsparsescenerepresentation}  & IL & 6.38	& 0.00	& 10.02\\
    \midrule
    ORION~\cite{fu2025orion} & IL+RL & 11.13 & 0.00 & 15.98 \\
    InternVL2-1B (SimLingo)~\cite{renz2025simlingo}  & IL & 13.11 & 0.00 & 19.55 \\
    InternVL3-1B~\cite{zhu2025internvl3}  & IL & 18.43 & 0.00 & 21.17\\
    \midrule
    \ours (InternVL2-1B) & IL+RL & \textbf{37.44} & \textbf{25.00} & \textbf{41.91} \\
    \ours (InternVL3-1B) & IL+RL & \textbf{39.89} & \textbf{25.00} & \textbf{47.24}

\\
    \bottomrule
  \end{tabular}
  }
  \vspace{-3pt}
  \label{tab:longest6}
\end{table*}

\subsection{Benchmarks}
\vspace{-3pt}
\noindent\textbf{CARLA longest6 v2.} The CARLA longest6 v2 benchmark consists of 36 routes from Town01-06 with an average distance of 1-2 kilometers, featuring 5-21 pre-crash, safety critical scenarios of 7 different types~\cite{carlaleaderboard,jaeger2025carl}. It is a challenging benchmark due to long routes, fast background traffic and complex scenarios.

\noindent\textbf{Bench2Drive.} Bench2Drive~\cite{jia2024bench2drive} is a CARLA-based benchmark that has 220 driving routes from Town01-15, each featuring a distinct, safety-critical scenario, with an average route length of 150 meters.

\noindent\textbf{nuScenes.} nuScenes~\cite{caesar2020nuscenes} consists of 1,000 scenes captured by a synchronized camera array of 6 cameras. Each scene lasts about 20 seconds at 2 fps. The dataset is split into 700 scenes for training, 150 scenes for validation and 150 scenes for testing. 

\subsection{Results}

In this section, we report results of our \ours evaluated on various benchmarks.

\subsubsection{CARLA longest6 v2}

We summarize the results on the CARLA longest6 v2 benchmark in Table~\ref{tab:longest6}. It is evident that even with large-scale imitation learning, all previous methods perform poorly with a success rate of 0\% (failing every route), due to long driving distances and complex scenarios. But after we conduct RL finetuning, \ours improves the performance of the VLA policies across the board, with significant jumps under all metrics, leading to a much more capable driver. This validates the effectiveness of the post-training RL framework in our \ours. We provide qualitative predictions of \ours in Figure~\ref{fig:demo-clear}.

\begin{table*}[t]
\centering
\caption{\small Comparison of closed-loop planning and multi-ability performance on the Bench2Drive benchmark.}
\label{tab:b2d}
\vspace{-5pt}
\resizebox{0.98\textwidth}{!}{
\begin{tabular}{lp{0.9cm}ccccccccccc}
\toprule
\multirow{2}{*}{Method} & 
\multicolumn{4}{c}{Closed-Loop Metrics} &
\multicolumn{6}{c}{Multi-Ability Test (\%) $\uparrow$} \\
\cmidrule(lr){2-5}
\cmidrule(lr){6-11}
 & 
DS$\uparrow$ &
SR (\%)$\uparrow$ &
Efficiency$\uparrow$ &
Comfort$\uparrow$ &

Merging &
Overtaking &
E-Brake &
Give Way &
T.Sign &
Mean \\

\midrule
TCP-traj~\cite{wu2022trajectory}  &

59.9 & 30.0 & 76.5 & 18.1 & 
28.8 & 24.3 & 51.7 & 40.0 & 46.3 & 34.2 \\

% TCP-traj w/o distillation~\cite{wu2022trajectory} &

% 49.3 & 20.5 & 78.8 & 22.9 &
% 28.7 & 28.7 & 48.3 & 40.0 & 28.7 & 34.1 \\

ThinkTwice~\cite{jia2023think} & 

62.4 & 31.2 & 69.3 & 16.2 &
27.4 & 18.4 & 35.8 & 50.0 & 54.2 & 37.2 \\

DriveAdapter~\cite{jia2023driveadapter} &

64.2 & 33.1 & 70.2 & 16.0 & 
28.4 & 28.4 & 47.5 & 50.0 & 56.4 & 42.1 \\

UniAD-Base~\cite{hu2023planning} &

45.8 & 16.4 & 129.2 & 43.6 & 
8.9 & 9.3 & 20.0 & 20.0 & 14.2 & 14.5 \\

VAD~\cite{jiang2023vad} &

42.4 & 15.0 & 157.9 & 46.0 & 
11.4 & 11.4 & 18.6 & 20.0 & 19.2 & 18.1 \\

%GenAD~\cite{zheng2024genad} &
%NC &
%44.81 & 15.90 & - & - & - &
%- & - & - & - & - & - \\

MomAD~\cite{song2025don} &

44.5 & 16.7 & 170.2 & \textbf{48.6} & 
- & - & - & - & - & - \\

DriveTransformer-L~\cite{jia2025drivetransformer} &

63.5 & 35.0 & 100.6 & 20.8 & 
17.6 & 35.0 & 48.4 & 40.0 & 52.1 & 38.6 \\

HiP-AD~\cite{tang2025hip}$\dagger$ &

\textbf{86.8} & 69.1 & 203.1 & 19.4 & 
50.0 & 84.4 & 83.3 & 40.0 & 50.5 & 65.9 \\

Qwen2.5~\cite{Bai2025qwen25}  &

63.9 & 31.6 & 119.3 & 10.1 &
14.3 & 28.9 & 30.1 & 30.0 & 24.7 & 25.6 \\

ORION~\cite{fu2025orion} &

77.7 & 54.6 & 151.5 & 17.4 & 
25.0 & 71.1 & 78.3 & 30.0 & 69.2 & 54.7 \\

ReCogDrive~\cite{li2026recogdrive} &

71.4 & 45.5 & 138.2 & 17.5 & 
29.7 & 20.0 & 69.1 & 20.0 & 71.3 & 42.0 \\

%DiffRefiner~\cite{yin2026diffrefiner}  &

%87.1 & 71.4 & - & - & 
%\textbf{63.8} & 60.0 & 85.0 & 50.0 & \textbf{86.3} & 69.0 \\

GeRo~\cite{yasarla2026generative} &

81.9 & 60.1 & 176.5 & 40.2 & 
40.1 & \textbf{78.2} & 87.3 & 50.0 & 76.8 & 66.5 \\

SimLingo~\cite{renz2025simlingo} &

85.1 & 67.3 & 259.2 & 33.7 & 
54.0 & 57.0 & \textbf{88.3} & 53.3 & \textbf{82.5} & 67.0 \\

Raw2Drive~\cite{yang2026raw2drive} & 71.4 & 50.2 & 214.2 & 22.42 & 43.4 & 51.1 & 60.0 & 50.0 & 62.3 & 53.3 \\

\midrule
\ours (InternVL2-1B) & 85.9	& 68.2 & 268.2 & 31.1	& 56.8	& 66.0	& 83.4	& 69.0	& 67.9 & 68.6 \\
\ours (InternVL3-1B) & \textbf{86.8} & \textbf{69.5} & \textbf{275.4}	&25.7 & \textbf{57.5} & 66.7	& 85.0	& \textbf{70.0}	& 69.8 & \textbf{69.8} \\
\bottomrule
\end{tabular}
}
\vspace{-5pt}
\end{table*}

\begin{table}[t]
\centering
\caption{\small Comparison of E2E-AD methods on the nuScenes~\cite{caesar2020nuscenes} validation set. $\dagger$ indicates zero-shot evaluated on nuScenes. M = Map, B = Bounding boxes, D = Depth, Mo = Motion prediction, O = Occupancy.}
%\vspace{-0.8em}
\label{tab:nusc}
\resizebox{0.85\columnwidth}{!}{
\begin{tabular}{ll|cccc|cccc}
\toprule
\multirow{2}{*}{Method} & \multirow{2}{*}{\makecell{Aux. Sup.}} & \multicolumn{4}{c|}{L2 ($\downarrow$)} & \multicolumn{4}{c}{Collision Rate ($\downarrow$)} \\
\cmidrule(lr){3-6} \cmidrule(lr){7-10}
& &  1s & 2s & 3s & Avg. & 1s & 2s & 3s & Avg. \\
\midrule
% IL~\cite{ratliff2006maximum}   & None & 0.44 & 1.15 & 2.47 & 1.35 & 0.08 & 0.27 & 1.95 & 0.77 & - \\
% NMP~\cite{zeng2019end}  & B+Mo & 0.53 & 1.25 & 2.67 & 1.48 & 0.04 & 0.12 & 0.87 & 0.34 & - \\
% FF~\cite{hu2021safe}  & FS & 0.55 & 1.20 & 2.54 & 1.43 & 0.06 & 0.17 & 1.07 & 0.43 & - \\
% EO~\cite{khurana2022differentiable}  & FS & 0.67 & 1.36 & 2.78 & 1.60 & 0.04 & 0.09 & 0.88 & 0.33 & - \\
% \hline
ST-P3~\cite{hu2022st}  & M+B+D & 1.33 & 2.11 & 2.90 & 2.11 & 0.23 & 0.62 & 1.27 & 0.71  \\
UniAD~\cite{hu2023planning}  & M+B+Mo+O  & 0.48 & 0.96 & 1.65 & 1.03 & 0.05 & 0.17 & 0.71 & 0.31 \\
OccNet~\cite{tong2023scene}  & M+B+O & 1.29 & 2.13 & 2.99 & 2.14 & 0.21 & 0.59 & 1.37 & 0.72  \\
% VAD-Tiny~\cite{jiang2023vad}  & M+B+Mo & 0.60 & 1.23 & 2.06 & 1.30 & 0.31 & 0.53 & 1.33 & 0.72 & 6.9$^\ddagger$ \\
VAD-Base~\cite{jiang2023vad}  & M+B+Mo & 0.54 & 1.15 & 1.98 & 1.22 & \textbf{0.04} & 0.39 & 1.17 & 0.53 \\
GenAD~\cite{zheng2024genad} & M+B+Mo & 0.36 & 0.83 & 1.55 & 0.91 & 0.06 & 0.23 & 1.00 & 0.43 \\
Senna~\cite{jiang2024senna} & M+B+Mo & 0.37 & 0.54 & 0.86 & 0.59 & 0.09 & \textbf{0.12} & \textbf{0.33} & \textbf{0.18} \\

Qwen2.5~\cite{Bai2025qwen25} & B+Mo & 0.52 & 0.80 & 1.56 & 0.96 & 0.10 & 0.37 & 1.32 & 0.60 \\
ORION~\cite{fu2025orion} & M+B+Mo & 0.43 & 0.64 & 1.01 & 0.69 & 0.09 & 0.29 & 0.94 & 0.44  \\
\hline
\rule{0pt}{1.2em}\ours (InternVL3-1B)$\dagger$ & N/A & \textbf{0.29} & \textbf{0.43} & \textbf{0.75} & \textbf{0.49} & 0.11 & 0.23 & 0.55 & 0.30\\
\bottomrule
\end{tabular}
}
\vspace{-8pt}
\end{table}

\subsubsection{Bench2Drive}

The results on the Bench2Drive benchmark is summarized in Table.~\ref{tab:b2d}. One can see \ours achieves the highest driving score and success rate, while also leading in driving efficiency and the averaged multi-ability score, bringing solid improvements on top of strong baselines. This proves the effectiveness of our RL finetuning design in \ours.

\subsubsection{nuScenes}

We further demonstrate \ours's driving capability on real-world driving datasets. The results are summarized in Table~\ref{tab:nusc}. We zero-shot evaluate \ours on the nuScenes validation set while the other models were trained on nuScenes. We can see that \ours obtains the best average $L_2$ error compared to previous methods and performs competitively under the collision rate metric, although our model has not seen any real-world data in training.

\begin{table*}[t]
  \small
  \centering
  \caption{\small Ablation study on RL action space.}
  \vspace{-5pt}
  \resizebox{0.65\textwidth}{!}{
  \begin{tabular}{ccccc}
    \toprule
    Method & Action Space & DS$\uparrow$ & SR (\%)$\uparrow$ & RC (\%)$\uparrow$ \\
    \midrule
    \multirow{2}{*}{\makecell{\ours (InternVL3-1B)\\}} & control & 33.91 & 19.44 & 39.23\\ 
     & waypoint & 39.89 & 25.00 & 47.24\\
    \bottomrule
  \end{tabular}
  }
  %\vspace{-3pt}
  \label{tab:action}
\end{table*}

\subsection{Ablation Studies}

We perform ablation studies on various aspects of \ours in this section. All experiments are done the CARLA longest6 v2~\cite{jaeger2025carl,carlaleaderboard} benchmark.

\textbf{RL Action Space.} We investigate how different output representation for RL affects the final planning performance. We compare with a low-level control policy, where we replace the waypoint head in the pretrained VLA with a small action head, which outputs driving actions (steering, merged throttle and brake), and trained with the same configuration with the rest of \ours. The results are summarized in Table~\ref{tab:action}. One can see a direct control policy experienced a big drop in planning performance. We posit that such direct control policy is not effectively utilizing the pretrained VLA knowledge by discarding the waypoints, whereas our residual waypoint formulation does.  

\textbf{Effect of Scaling.} We study the effects of increasing the mini-batch size in PPO. The results are summarized in Table~\ref{tab:scaling}. When using a small mini-batch size of 256, the training failed to converge. When we increase mini-batch size all the way to 4096 and scales to 100M samples, \ours achieves progressively better results, highlighting the importance of RL at scale.

\begin{table*}[t]
  \small
  \centering
  \caption{\small Ablation study on the effect of scaling.}
  \vspace{-5pt}
  \resizebox{0.8\textwidth}{!}{
  \begin{tabular}{ccccccc}
    \toprule
    Method & Samples & Total-batch & Mini-batch & DS$\uparrow$ & SR (\%)$\uparrow$ & RC (\%)$\uparrow$ \\
    \midrule
    \multirow{3}{*}{\makecell{\ours (InternVL3-1B)\\}} & 10M & 1024 & 256 & 12.72 & 0.0	& 17.44\\
    & 40M & 4096 & 1024 & 25.31 & 13.89 & 32.75\\
    & 100M & 16384 & 4096 & 39.89 & 25.00 & 47.24 \\
    \bottomrule
  \end{tabular}
  }
  %\vspace{-3pt}
  \label{tab:scaling}
\end{table*}

%% file: secs/5_limitation.tex
\section{Limitations}
\label{sec:limit}

We list a few limitations, which can be considered as part of future work:

\begin{itemize}
    \item Our method \ours sets new SotA for VLA policies using RL in a driving simulator. To be able to deploy it to the real world, Sim2Real~\cite{lin2025sim,zhao2020sim} transfer needs to be investigated. 
    \item Our current scaling approach is limited by the number of parallel CARLA simulation environments we can spawn at training time. Investigating to improve the stability of CARLA's GPU rendering stack, so that more servers can be reliably launched on a single machine, would be helpful for further scaling up RL.
    %\item We train \ours using the routes generated by~\cite{jaeger2025carl,carlaleaderboard}, where the CARLA leaderboard 2.0~\cite{carlaleaderboard} offers more types of scenarios, which we have not looked into.
    %\item Bounded by compute resources, we froze the vision encoder and LLM inside the VLA when carrying out RL finetuning. Investigating efficient schemes to unfreeze the model components can lead to observations as to how it affects planning performance. 
    \item We utilize the reward from~\cite{jaeger2025carl}, whose global optimum is same as the global optimum of the metric. When the metric does not possess a global optimum, how to design a good reward for VLA policies is an important venue to investigate. 

\end{itemize}

%% file: secs/6_conclusion.tex
\section{Conclusion}
\label{sec:conclusion}

In this work, we present \ours, a system that enables closed-loop RL at scale for end-to-end VLA driving policies. We start with large-scale open-loop pretraining with expert trajectories, equipping the policy with baseline driving capability. Then, we carry out RL by learning a novel residual policy around the pretrained waypoints prior. To enable effective scaling, we design a heterogeneous pipeline where we separate the simulation environment and the VLA learner, resolving resource contention between the two. We scale to 100M samples for PPO updates with a total-batch/mini-batch of 16384/4096, outperforming all previous methods with significant gains across the board on the CARLA longest6 v2 benchmark, while also maintaining competitive leads on the Bench2Drive and nuScenes benchmark.

%% file: main.bib
@String(ICLR = {Int. Conf. Learn. Represent.})

@String(ICLR  = {ICLR})

@article{chen2024end,
  title={End-to-end autonomous driving: Challenges and frontiers},
  author={Chen, Li and Wu, Penghao and Chitta, Kashyap and Jaeger, Bernhard and Geiger, Andreas and Li, Hongyang},
  journal={IEEE Transactions on Pattern Analysis and Machine Intelligence},
  volume={46},
  number={12},
  pages={10164--10183},
  year={2024},
  publisher={IEEE}
}

@inproceedings{hu2023planning,
  title={Planning-oriented autonomous driving},
  author={Hu, Yihan and Yang, Jiazhi and Chen, Li and Li, Keyu and Sima, Chonghao and Zhu, Xizhou and Chai, Siqi and Du, Senyao and Lin, Tianwei and Wang, Wenhai and others},
  booktitle={Proceedings of the IEEE/CVF conference on computer vision and pattern recognition},
  pages={17853--17862},
  year={2023}
}

@inproceedings{jiang2023vad,
  title={Vad: Vectorized scene representation for efficient autonomous driving},
  author={Jiang, Bo and Chen, Shaoyu and Xu, Qing and Liao, Bencheng and Chen, Jiajie and Zhou, Helong and Zhang, Qian and Liu, Wenyu and Huang, Chang and Wang, Xinggang},
  booktitle={Proceedings of the IEEE/CVF International Conference on Computer Vision},
  pages={8340--8350},
  year={2023}
}

@article{chen2024vadv2,
  title={Vadv2: End-to-end vectorized autonomous driving via probabilistic planning},
  author={Chen, Shaoyu and Jiang, Bo and Gao, Hao and Liao, Bencheng and Xu, Qing and Zhang, Qian and Huang, Chang and Liu, Wenyu and Wang, Xinggang},
  journal={arXiv preprint arXiv:2402.13243},
  year={2024}
}

@inproceedings{li2025navigationguidedsparsescenerepresentation,
  title={Navigation-Guided Sparse Scene Representation for End-to-End Autonomous Driving},
  author={Peidong Li and Dixiao Cui},
  booktitle={International Conference on Learning Representations (ICLR)},
  year={2025}
}

@inproceedings{jia2023driveadapter,
  title={Driveadapter: Breaking the coupling barrier of perception and planning in end-to-end autonomous driving},
  author={Jia, Xiaosong and Gao, Yulu and Chen, Li and Yan, Junchi and Liu, Patrick Langechuan and Li, Hongyang},
  booktitle={Proceedings of the IEEE/CVF International Conference on Computer Vision},
  pages={7953--7963},
  year={2023}
}

@inproceedings{chen2024internvl,
  title={Internvl: Scaling up vision foundation models and aligning for generic visual-linguistic tasks},
  author={Chen, Zhe and Wu, Jiannan and Wang, Wenhai and Su, Weijie and Chen, Guo and Xing, Sen and Zhong, Muyan and Zhang, Qinglong and Zhu, Xizhou and Lu, Lewei and others},
  booktitle={Proceedings of the IEEE/CVF conference on computer vision and pattern recognition},
  pages={24185--24198},
  year={2024}
}

@article{qwen,
  title={Qwen Technical Report},
  author={Jinze Bai and Shuai Bai and Yunfei Chu and Zeyu Cui and Kai Dang and Xiaodong Deng and Yang Fan and Wenbin Ge and Yu Han and Fei Huang and Binyuan Hui and Luo Ji and Mei Li and Junyang Lin and Runji Lin and Dayiheng Liu and Gao Liu and Chengqiang Lu and Keming Lu and Jianxin Ma and Rui Men and Xingzhang Ren and Xuancheng Ren and Chuanqi Tan and Sinan Tan and Jianhong Tu and Peng Wang and Shijie Wang and Wei Wang and Shengguang Wu and Benfeng Xu and Jin Xu and An Yang and Hao Yang and Jian Yang and Shusheng Yang and Yang Yao and Bowen Yu and Hongyi Yuan and Zheng Yuan and Jianwei Zhang and Xingxuan Zhang and Yichang Zhang and Zhenru Zhang and Chang Zhou and Jingren Zhou and Xiaohuan Zhou and Tianhang Zhu},
  journal={arXiv preprint arXiv:2309.16609},
  year={2023}
}

@article{achiam2023gpt,
  title={Gpt-4 technical report},
  author={Achiam, Josh and Adler, Steven and Agarwal, Sandhini and Ahmad, Lama and Akkaya, Ilge and Aleman, Florencia Leoni and Almeida, Diogo and Altenschmidt, Janko and Altman, Sam and Anadkat, Shyamal and others},
  journal={arXiv preprint arXiv:2303.08774},
  year={2023}
}

@article{singh2025openai,
  title={Openai gpt-5 system card},
  author={Singh, Aaditya and Fry, Adam and Perelman, Adam and Tart, Adam and Ganesh, Adi and El-Kishky, Ahmed and McLaughlin, Aidan and Low, Aiden and Ostrow, AJ and Ananthram, Akhila and others},
  journal={arXiv preprint arXiv:2601.03267},
  year={2025}
}

@inproceedings{caesar2020nuscenes,
  title={nuscenes: A multimodal dataset for autonomous driving},
  author={Caesar, Holger and Bankiti, Varun and Lang, Alex H and Vora, Sourabh and Liong, Venice Erin and Xu, Qiang and Krishnan, Anush and Pan, Yu and Baldan, Giancarlo and Beijbom, Oscar},
  booktitle={Proceedings of the IEEE/CVF conference on computer vision and pattern recognition},
  pages={11621--11631},
  year={2020}
}

@article{caesar2021nuplan,
  title={nuplan: A closed-loop ml-based planning benchmark for autonomous vehicles},
  author={Caesar, Holger and Kabzan, Juraj and Tan, Kok Seang and Fong, Whye Kit and Wolff, Eric and Lang, Alex and Fletcher, Luke and Beijbom, Oscar and Omari, Sammy},
  journal={arXiv preprint arXiv:2106.11810},
  year={2021}
}

@inproceedings{sun2020scalability,
  title={Scalability in perception for autonomous driving: Waymo open dataset},
  author={Sun, Pei and Kretzschmar, Henrik and Dotiwalla, Xerxes and Chouard, Aurelien and Patnaik, Vijaysai and Tsui, Paul and Guo, James and Zhou, Yin and Chai, Yuning and Caine, Benjamin and others},
  booktitle={Proceedings of the IEEE/CVF conference on computer vision and pattern recognition},
  pages={2446--2454},
  year={2020}
}

@article{dauner2024navsim,
  title={Navsim: Data-driven non-reactive autonomous vehicle simulation and benchmarking},
  author={Dauner, Daniel and Hallgarten, Marcel and Li, Tianyu and Weng, Xinshuo and Huang, Zhiyu and Yang, Zetong and Li, Hongyang and Gilitschenski, Igor and Ivanovic, Boris and Pavone, Marco and others},
  journal={Advances in Neural Information Processing Systems},
  volume={37},
  pages={28706--28719},
  year={2024}
}

@article{jia2024bench2drive,
  title={Bench2drive: Towards multi-ability benchmarking of closed-loop end-to-end autonomous driving},
  author={Jia, Xiaosong and Yang, Zhenjie and Li, Qifeng and Zhang, Zhiyuan and Yan, Junchi},
  journal={Advances in Neural Information Processing Systems},
  volume={37},
  pages={819--844},
  year={2024}
}

@article{chitta2022transfuser,
  title={Transfuser: Imitation with transformer-based sensor fusion for autonomous driving},
  author={Chitta, Kashyap and Prakash, Aditya and Jaeger, Bernhard and Yu, Zehao and Renz, Katrin and Geiger, Andreas},
  journal={IEEE transactions on pattern analysis and machine intelligence},
  volume={45},
  number={11},
  pages={12878--12895},
  year={2022},
  publisher={IEEE}
}

@article{hwang2024emma,
  title={Emma: End-to-end multimodal model for autonomous driving},
  author={Hwang, Jyh-Jing and Xu, Runsheng and Lin, Hubert and Hung, Wei-Chih and Ji, Jingwei and Choi, Kristy and Huang, Di and He, Tong and Covington, Paul and Sapp, Benjamin and others},
  journal={arXiv preprint arXiv:2410.23262},
  year={2024}
}

@inproceedings{xing2025openemma,
  title={Openemma: Open-source multimodal model for end-to-end autonomous driving},
  author={Xing, Shuo and Qian, Chengyuan and Wang, Yuping and Hua, Hongyuan and Tian, Kexin and Zhou, Yang and Tu, Zhengzhong},
  booktitle={Proceedings of the Winter Conference on Applications of Computer Vision},
  pages={1001--1009},
  year={2025}
}

@inproceedings{fu2025orion,
  title={Orion: A holistic end-to-end autonomous driving framework by vision-language instructed action generation},
  author={Fu, Haoyu and Zhang, Diankun and Zhao, Zongchuang and Cui, Jianfeng and Liang, Dingkang and Zhang, Chong and Zhang, Dingyuan and Xie, Hongwei and Wang, Bing and Bai, Xiang},
  booktitle={Proceedings of the IEEE/CVF International Conference on Computer Vision},
  pages={24823--24834},
  year={2025}
}

@inproceedings{renz2025simlingo,
  title={Simlingo: Vision-only closed-loop autonomous driving with language-action alignment},
  author={Renz, Katrin and Chen, Long and Arani, Elahe and Sinavski, Oleg},
  booktitle={Proceedings of the Computer Vision and Pattern Recognition Conference},
  pages={11993--12003},
  year={2025}
}

@inproceedings{chitta2021neat,
  title={Neat: Neural attention fields for end-to-end autonomous driving},
  author={Chitta, Kashyap and Prakash, Aditya and Geiger, Andreas},
  booktitle={Proceedings of the IEEE/CVF International Conference on Computer Vision},
  pages={15793--15803},
  year={2021}
}

@inproceedings{hawke2020urban,
  title={Urban driving with conditional imitation learning},
  author={Hawke, Jeffrey and Shen, Richard and Gurau, Corina and Sharma, Siddharth and Reda, Daniele and Nikolov, Nikolay and Mazur, Przemys{\l}aw and Micklethwaite, Sean and Griffiths, Nicolas and Shah, Amar and others},
  booktitle={2020 IEEE International Conference on Robotics and Automation (ICRA)},
  pages={251--257},
  year={2020},
  organization={IEEE}
}

@inproceedings{jia2023think,
  title={Think twice before driving: Towards scalable decoders for end-to-end autonomous driving},
  author={Jia, Xiaosong and Wu, Penghao and Chen, Li and Xie, Jiangwei and He, Conghui and Yan, Junchi and Li, Hongyang},
  booktitle={Proceedings of the IEEE/CVF Conference on Computer Vision and Pattern Recognition},
  pages={21983--21994},
  year={2023}
}

@inproceedings{shao2023safety,
  title={Safety-enhanced autonomous driving using interpretable sensor fusion transformer},
  author={Shao, Hao and Wang, Letian and Chen, Ruobing and Li, Hongsheng and Liu, Yu},
  booktitle={Conference on Robot Learning},
  pages={726--737},
  year={2023},
  organization={PMLR}
}

@article{team2023gemini,
  title={Gemini: a family of highly capable multimodal models},
  author={Team, Gemini and Anil, Rohan and Borgeaud, Sebastian and Alayrac, Jean-Baptiste and Yu, Jiahui and Soricut, Radu and Schalkwyk, Johan and Dai, Andrew M and Hauth, Anja and Millican, Katie and others},
  journal={arXiv preprint arXiv:2312.11805},
  year={2023}
}

@article{comanici2025gemini,
  title={Gemini 2.5: Pushing the frontier with advanced reasoning, multimodality, long context, and next generation agentic capabilities},
  author={Comanici, Gheorghe and Bieber, Eric and Schaekermann, Mike and Pasupat, Ice and Sachdeva, Noveen and Dhillon, Inderjit and Blistein, Marcel and Ram, Ori and Zhang, Dan and Rosen, Evan and others},
  journal={arXiv preprint arXiv:2507.06261},
  year={2025}
}

@article{cao2025pseudo,
  title={Pseudo-simulation for autonomous driving},
  author={Cao, Wei and Hallgarten, Marcel and Li, Tianyu and Dauner, Daniel and Gu, Xunjiang and Wang, Caojun and Miron, Yakov and Aiello, Marco and Li, Hongyang and Gilitschenski, Igor and others},
  journal={Conference on Robot Learning},
  year={2025}
}

@article{wu2022trajectory,
  title={Trajectory-guided control prediction for end-to-end autonomous driving: A simple yet strong baseline},
  author={Wu, Penghao and Jia, Xiaosong and Chen, Li and Yan, Junchi and Li, Hongyang and Qiao, Yu},
  journal={Advances in Neural Information Processing Systems},
  volume={35},
  pages={6119--6132},
  year={2022}
}

@inproceedings{li2024think2drive,
  title={Think2drive: Efficient reinforcement learning by thinking with latent world model for autonomous driving (in carla-v2)},
  author={Li, Qifeng and Jia, Xiaosong and Wang, Shaobo and Yan, Junchi},
  booktitle={European conference on computer vision},
  pages={142--158},
  year={2024},
  organization={Springer}
}

@article{jaeger2025carl,
  title={Carl: Learning scalable planning policies with simple rewards},
  author={Jaeger, Bernhard and Dauner, Daniel and Bei{\ss}wenger, Jens and Gerstenecker, Simon and Chitta, Kashyap and Geiger, Andreas},
  journal={arXiv preprint arXiv:2504.17838},
  year={2025}
}

@book{sutton1998reinforcement,
  title={Reinforcement learning: An introduction},
  author={Sutton, Richard S and Barto, Andrew G and others},
  volume={1},
  number={1},
  year={1998},
  publisher={MIT press Cambridge}
}

@article{renz2022plant,
  title={Plant: Explainable planning transformers via object-level representations},
  author={Renz, Katrin and Chitta, Kashyap and Mercea, Otniel-Bogdan and Koepke, A and Akata, Zeynep and Geiger, Andreas},
  journal={Conference on Robotic Learning},
  year={2022}
}

@article{schulman2017proximal,
  title={Proximal policy optimization algorithms},
  author={Schulman, John and Wolski, Filip and Dhariwal, Prafulla and Radford, Alec and Klimov, Oleg},
  journal={arXiv preprint arXiv:1707.06347},
  year={2017}
}

@article{zhu2025internvl3,
  title={Internvl3: Exploring advanced training and test-time recipes for open-source multimodal models},
  author={Zhu, Jinguo and Wang, Weiyun and Chen, Zhe and Liu, Zhaoyang and Ye, Shenglong and Gu, Lixin and Tian, Hao and Duan, Yuchen and Su, Weijie and Shao, Jie and others},
  journal={arXiv preprint arXiv:2504.10479},
  year={2025}
}

@article{Bai2025qwen25,
  title={Qwen2.5-VL Technical Report},
  author={Bai, Shuai and Chen, Keqin and Liu, Xuejing and Wang, Jialin and Ge, Wenbin and Song, Sibo and Dang Kai and Wang, Peng and Wang Shijie and Tang Jun and Zhong Humen and Zhu, Yuanzhi and Yang, Mingkun and Li, Zhaohai and Wan, Jianqiang and Wang, Pengfei and Ding, Wei and Fu, Zheren and Xu, Yiheng and Ye, Jiabo and Zhang Xi and Xie Tianbao and Cheng, Zesen and Zhang, Hang  and Yang, Zhibo and Xu, Haiyang and Lin, Junyang},
  journal={arXiv preprint arXiv:2502.13923},
  year={2025}
}

@inproceedings{dosovitskiy2017carla,
  title={CARLA: An open urban driving simulator},
  author={Dosovitskiy, Alexey and Ros, German and Codevilla, Felipe and Lopez, Antonio and Koltun, Vladlen},
  booktitle={Conference on robot learning},
  pages={1--16},
  year={2017},
  organization={PMLR}
}

@article{carlaleaderboard,
  title={The CARLA leaderboard 2.0},
  author={The CARLA team},
  journal={https://leaderboard.carla.org},
  year={2022}
}

@article{wijmans2019dd,
  title={Dd-ppo: Learning near-perfect pointgoal navigators from 2.5 billion frames},
  author={Wijmans, Erik and Kadian, Abhishek and Morcos, Ari and Lee, Stefan and Essa, Irfan and Parikh, Devi and Savva, Manolis and Batra, Dhruv},
  journal={International Conference on Learning Representations (ICLR)},
  year={2020}
}

@article{carlaissues,
  title={The CARLA issues section},
  author={The CARLA community},
  journal={https://github.com/carla-simulator/carla/issues},
  year={2022}
}

@misc{rfc4254,
  author       = {Tatu Ylonen and Chris Lonvick},
  title        = {The Secure Shell (SSH) Connection Protocol},
  howpublished = {RFC 4254 (Standards Track)},
  year         = {2006},
  month        = jan,
  publisher    = {RFC Editor},
  doi          = {10.17487/RFC4254},
  url          = {https://www.rfc-editor.org/rfc/rfc4254}
}

@inproceedings{zheng2024genad,
  title={Genad: Generative end-to-end autonomous driving},
  author={Zheng, Wenzhao and Song, Ruiqi and Guo, Xianda and Zhang, Chenming and Chen, Long},
  booktitle={European Conference on Computer Vision},
  pages={87--104},
  year={2024},
  organization={Springer}
}

@article{jia2025drivetransformer,
  title={Drivetransformer: Unified transformer for scalable end-to-end autonomous driving},
  author={Jia, Xiaosong and You, Junqi and Zhang, Zhiyuan and Yan, Junchi},
  journal={arXiv preprint arXiv:2503.07656},
  year={2025}
}

@article{engine2018unreal,
  title={Unreal engine},
  author={Engine, Unreal},
  journal={Retrieved from Unreal Engine: https://www. unrealengine. com/en-US/what-is-unreal-engine-4},
  year={2018}
}

@article{loshchilov2017decoupled,
  title={Decoupled weight decay regularization},
  author={Loshchilov, Ilya and Hutter, Frank},
  journal={arXiv preprint arXiv:1711.05101},
  year={2017}
}

@article{hu2022lora,
  title={Lora: Low-rank adaptation of large language models.},
  author={Hu, Edward J and Shen, Yelong and Wallis, Phillip and Allen-Zhu, Zeyuan and Li, Yuanzhi and Wang, Shean and Wang, Liang and Chen, Weizhu and others},
  journal={Iclr},
  volume={1},
  number={2},
  pages={3},
  year={2022}
}

@inproceedings{song2025don,
  title={Don't shake the wheel: Momentum-aware planning in end-to-end autonomous driving},
  author={Song, Ziying and Jia, Caiyan and Liu, Lin and Pan, Hongyu and Zhang, Yongchang and Wang, Junming and Zhang, Xingyu and Xu, Shaoqing and Yang, Lei and Luo, Yadan},
  booktitle={Proceedings of the IEEE/CVF Conference on Computer Vision and Pattern Recognition},
  pages={22432--22441},
  year={2025}
}

@inproceedings{tang2025hip,
  title={Hip-ad: Hierarchical and multi-granularity planning with deformable attention for autonomous driving in a single decoder},
  author={Tang, Yingqi and Xu, Zhuoran and Meng, Zhaotie and Cheng, Erkang},
  booktitle={Proceedings of the IEEE/CVF International Conference on Computer Vision},
  pages={25605--25615},
  year={2025}
}

@inproceedings{li2026recogdrive,
  title     = {ReCogDrive: A Reinforced Cognitive Framework for End-to-End Autonomous Driving},
  author    = {Li, Yongkang and Xiong, Kaixin and Guo, Xiangyu and Li, Fang and Yan, Sixu and Xu, Gangwei and Zhou, Lijun and Chen, Long and Sun, Haiyang and Wang, Bing and others},
  booktitle = {International Conference on Learning Representations (ICLR)},
  year      = {2026}
}

@article{yasarla2026generative,
  title={Generative Scenario Rollouts for End-to-End Autonomous Driving},
  author={Yasarla, Rajeev and Hegde, Deepti and Han, Shizhong and Cheng, Hsin-Pai and Shi, Yunxiao and Sadeghigooghari, Meysam and Mahajan, Shweta and Bhattacharyya, Apratim and Liu, Litian and Garrepalli, Risheek and others},
  journal={arXiv preprint arXiv:2601.11475},
  year={2026}
}

@article{jiang2024senna,
  title={Senna: Bridging large vision-language models and end-to-end autonomous driving},
  author={Jiang, Bo and Chen, Shaoyu and Liao, Bencheng and Zhang, Xingyu and Yin, Wei and Zhang, Qian and Huang, Chang and Liu, Wenyu and Wang, Xinggang},
  journal={arXiv preprint arXiv:2410.22313},
  year={2024}
}

@inproceedings{tong2023scene,
  title={Scene as occupancy},
  author={Tong, Wenwen and Sima, Chonghao and Wang, Tai and Chen, Li and Wu, Silei and Deng, Hanming and Gu, Yi and Lu, Lewei and Luo, Ping and Lin, Dahua and others},
  booktitle={Proceedings of the IEEE/CVF International Conference on Computer Vision},
  pages={8406--8415},
  year={2023}
}

@inproceedings{hu2022st,
  title={St-p3: End-to-end vision-based autonomous driving via spatial-temporal feature learning},
  author={Hu, Shengchao and Chen, Li and Wu, Penghao and Li, Hongyang and Yan, Junchi and Tao, Dacheng},
  booktitle={European Conference on Computer Vision},
  pages={533--549},
  year={2022},
  organization={Springer}
}

@article{gao2026rad,
  title={Rad: Training an end-to-end driving policy via large-scale 3dgs-based reinforcement learning},
  author={Gao, Hao and Chen, Shaoyu and Jiang, Bo and Liao, Bencheng and Shi, Yiang and Guo, Xiaoyang and Pu, Yuechuan and Li, Xiangyu and Liu, Wenyu and Zhang, Qian and others},
  journal={Advances in Neural Information Processing Systems},
  volume={38},
  pages={32551--32576},
  year={2026}
}

@inproceedings{huang2025gen,
  title={Gen-drive: Enhancing diffusion generative driving policies with reward modeling and reinforcement learning fine-tuning},
  author={Huang, Zhiyu and Weng, Xinshuo and Igl, Maximilian and Chen, Yuxiao and Cao, Yulong and Ivanovic, Boris and Pavone, Marco and Lv, Chen},
  booktitle={2025 IEEE International Conference on Robotics and Automation (ICRA)},
  pages={3445--3451},
  year={2025},
  organization={IEEE}
}

@article{li2025finetuning,
  title={Finetuning generative trajectory model with reinforcement learning from human feedback},
  author={Li, Derun and Ren, Jianwei and Wang, Yue and Wen, Xin and Li, Pengxiang and Xu, Leimeng and Zhan, Kun and Xia, Zhongpu and Jia, Peng and Lang, Xianpeng and others},
  journal={arXiv e-prints},
  pages={arXiv--2503},
  year={2025}
}

@article{ouyang2022training,
  title={Training language models to follow instructions with human feedback},
  author={Ouyang, Long and Wu, Jeffrey and Jiang, Xu and Almeida, Diogo and Wainwright, Carroll and Mishkin, Pamela and Zhang, Chong and Agarwal, Sandhini and Slama, Katarina and Ray, Alex and others},
  journal={Advances in neural information processing systems},
  volume={35},
  pages={27730--27744},
  year={2022}
}

@article{jiang2025alphadrive,
  title={Alphadrive: Unleashing the power of vlms in autonomous driving via reinforcement learning and reasoning},
  author={Jiang, Bo and Chen, Shaoyu and Zhang, Qian and Liu, Wenyu and Wang, Xinggang},
  journal={arXiv preprint arXiv:2503.07608},
  year={2025}
}

@article{zhou2026autovla,
  title={Autovla: A vision-language-action model for end-to-end autonomous driving with adaptive reasoning and reinforcement fine-tuning},
  author={Zhou, Zewei and Cai, Tianhui and Zhao, Seth and Zhang, Yun and Huang, Zhiyu and Zhou, Bolei and Ma, Jiaqi},
  journal={Advances in Neural Information Processing Systems},
  volume={38},
  pages={27920--27956},
  year={2026}
}

@article{shao2024deepseekmath,
  title={Deepseekmath: Pushing the limits of mathematical reasoning in open language models},
  author={Shao, Zhihong and Wang, Peiyi and Zhu, Qihao and Xu, Runxin and Song, Junxiao and Bi, Xiao and Zhang, Haowei and Zhang, Mingchuan and Li, YK and Wu, Yang and others},
  journal={arXiv preprint arXiv:2402.03300},
  year={2024}
}

@article{yang2026raw2drive,
  title={Raw2drive: Reinforcement learning with aligned world models for end-to-end autonomous driving (in carla v2)},
  author={Yang, Zhenjie and Jia, Xiaosong and Li, Qifeng and Yang, Xue and Yao, Maoqing and Yan, Junchi},
  journal={Advances in Neural Information Processing Systems},
  volume={38},
  pages={134122--134147},
  year={2026}
}

@article{lin2025sim,
  title={Sim-to-real reinforcement learning for vision-based dexterous manipulation on humanoids},
  author={Lin, Toru and Sachdev, Kartik and Fan, Linxi and Malik, Jitendra and Zhu, Yuke},
  journal={arXiv preprint arXiv:2502.20396},
  year={2025}
}

@inproceedings{zhao2020sim,
  title={Sim-to-real transfer in deep reinforcement learning for robotics: a survey},
  author={Zhao, Wenshuai and Queralta, Jorge Pe{\~n}a and Westerlund, Tomi},
  booktitle={2020 IEEE symposium series on computational intelligence (SSCI)},
  pages={737--744},
  year={2020},
  organization={IEEE}
}

@article{mao2023gpt,
  title={Gpt-driver: Learning to drive with gpt},
  author={Mao, Jiageng and Qian, Yuxi and Ye, Junjie and Zhao, Hang and Wang, Yue},
  journal={arXiv preprint arXiv:2310.01415},
  year={2023}
}

@inproceedings{fu2024drive,
  title={Drive like a human: Rethinking autonomous driving with large language models},
  author={Fu, Daocheng and Li, Xin and Wen, Licheng and Dou, Min and Cai, Pinlong and Shi, Botian and Qiao, Yu},
  booktitle={2024 IEEE/CVF Winter Conference on Applications of Computer Vision Workshops (WACVW)},
  pages={910--919},
  year={2024},
  organization={IEEE}
}

@article{huang2024drivemm,
  title={Drivemm: All-in-one large multimodal model for autonomous driving},
  author={Huang, Zhijian and Feng, Chengjian and Yan, Feng and Xiao, Baihui and Jie, Zequn and Zhong, Yujie and Liang, Xiaodan and Ma, Lin},
  journal={arXiv preprint arXiv:2412.07689},
  volume={2},
  number={3},
  pages={8},
  year={2024}
}

@inproceedings{liu2023mtd,
  title={Mtd-gpt: A multi-task decision-making gpt model for autonomous driving at unsignalized intersections},
  author={Liu, Jiaqi and Hang, Peng and Qi, Xiao and Wang, Jianqiang and Sun, Jian},
  booktitle={2023 IEEE 26th International Conference on Intelligent Transportation Systems (ITSC)},
  pages={5154--5161},
  year={2023},
  organization={IEEE}
}

@inproceedings{ma2024dolphins,
  title={Dolphins: Multimodal language model for driving},
  author={Ma, Yingzi and Cao, Yulong and Sun, Jiachen and Pavone, Marco and Xiao, Chaowei},
  booktitle={European Conference on Computer Vision},
  pages={403--420},
  year={2024},
  organization={Springer}
}

@inproceedings{paul2024lego,
  title={Lego-drive: Language-enhanced goal-oriented closed-loop end-to-end autonomous driving},
  author={Paul, Pranjal and Garg, Anant and Choudhary, Tushar and Singh, Arun Kumar and Krishna, K Madhava},
  booktitle={2024 IEEE/RSJ International Conference on Intelligent Robots and Systems (IROS)},
  pages={10020--10026},
  year={2024},
  organization={IEEE}
}

@inproceedings{wang2024drive,
  title={Drive anywhere: Generalizable end-to-end autonomous driving with multi-modal foundation models},
  author={Wang, Tsun-Hsuan and Maalouf, Alaa and Xiao, Wei and Ban, Yutong and Amini, Alexander and Rosman, Guy and Karaman, Sertac and Rus, Daniela},
  booktitle={2024 IEEE International Conference on Robotics and Automation (ICRA)},
  pages={6687--6694},
  year={2024},
  organization={IEEE}
}

@article{wang2023drivemlm,
  title={Drivemlm: Aligning multi-modal large language models with behavioral planning states for autonomous driving},
  author={Wang, Wenhai and Xie, Jiangwei and Hu, ChuanYang and Zou, Haoming and Fan, Jianan and Tong, Wenwen and Wen, Yang and Wu, Silei and Deng, Hanming and Li, Zhiqi and others},
  journal={arXiv preprint arXiv:2312.09245},
  year={2023}
}

@article{tian2024drivevlm,
  title={Drivevlm: The convergence of autonomous driving and large vision-language models},
  author={Tian, Xiaoyu and Gu, Junru and Li, Bailin and Liu, Yicheng and Wang, Yang and Zhao, Zhiyong and Zhan, Kun and Jia, Peng and Lang, Xianpeng and Zhao, Hang},
  journal={arXiv preprint arXiv:2402.12289},
  year={2024}
}

@inproceedings{jiang2025survey,
  title={A survey on vision-language-action models for autonomous driving},
  author={Jiang, Sicong and Huang, Zilin and Qian, Kangan and Luo, Ziang and Zhu, Tianze and Zhong, Yang and Tang, Yihong and Kong, Menglin and Wang, Yunlong and Jiao, Siwen and others},
  booktitle={Proceedings of the IEEE/CVF International Conference on Computer Vision},
  pages={4524--4536},
  year={2025}
}
